\documentclass[review,3p]{elsarticle}
\usepackage{CJKutf8}
\graphicspath{ {./figures/} }
\usepackage{hyperref}
\usepackage{float}
\usepackage{verbatim}
\graphicspath{{./}}
\usepackage{xcolor}
\usepackage{xpatch}
\usepackage{graphicx}
\usepackage[section]{placeins}
\usepackage{float}
\usepackage{amsthm,amsmath,amssymb}
\usepackage{mathrsfs}
\usepackage{amsmath}
\usepackage{subfigure}% 并排子图
\graphicspath{{./}}
\usepackage{amssymb}
\usepackage{float}
\usepackage{natbib}
\usepackage{tabu}                     % 表格插入
\usepackage{multirow}                 % 一般用以设计表格，将所行合并
\usepackage{multicol}                 % 合并多列
\usepackage{multirow}                % 合并多行
\usepackage{float}                    % 图片浮动
\usepackage{makecell}                 % 三线表-竖线
\usepackage{booktabs} 
\usepackage{caption}
\graphicspath{ {./figures/} }
\usepackage{hyperref}
\usepackage{float}
\usepackage{verbatim} %comments
\usepackage{apalike}
\restylefloat{figure}
\floatstyle{plaintop} %table caption at top
\restylefloat{table}

\begin{document}
\begin{CJK}{UTF8}{gbsn}
\begin{frontmatter}

\title{Robust image segmentation model based on binary level set}

\author[1]{Wenqi Zhao}
\ead{zwq@stu.cqu.edu.cn}
\author[2]{Jiacheng Sang}
\ead{sangjiacheng@stu.scu.edu.cn}

\author[1]{Yonglu Shu\corref{cor1}}
%\fnmark[1]
\cortext[cor1]{Corresponding author}
\ead{shuyonglu@cqu.edu.cn}
\author[1]{Dong Li}
\ead{lid@cqu.edu.cn}
\address[1]{College of Mathematics and Statistics, Chongqing University, 401331,  Chongqing, China}
\address[2]{State Key Laboratory of Biotherapy, Sichuan University,  610000,      Sichuan, China}

\begin{abstract}
In order to improve the robustness of traditional image segmentation models to noise, this paper models the illumination term in intensity inhomogeneity images. Additionally, to enhance the model's robustness to noisy images, we incorporate the binary level set model into the proposed model. Compared to the traditional level set, the binary level set eliminates the need for continuous reinitialization. Moreover, by introducing the variational operator GL, our model demonstrates better capability in segmenting noisy images. Finally, we employ the three-step splitting operator method for solving, and the effectiveness of the proposed model is demonstrated on various images.
\end{abstract}


\end{frontmatter}

\section{Introduction}
\label{introduction}

	With the continuous development of image processing technology, image segmentation plays a key role in various fields such as medical image analysis, 3D reconstruction, and object tracking. In particular, in the context of image segmentation-assisted medical diagnosis, images often suffer from varying degrees of in and noise due to imaging device limitations, which can impact the accuracy of medical diagnosis and treatment. Therefore, for medical images with severe intensity inhomogeneity and noise, a robust and accurate segmentation model is especially important for medical diagnosis.
Over the past few decades, numerous segmentation models have been proposed specifically for intensity inhomogeneity and noisy images. Li et al. \cite{4623242} proposed the Locally Based Fitting (LBF) model, which utilizes the intensity information of local image regions. Zhang \cite{7059203} modeled intensity inhomogeneity images as Gaussian distributions with different means and variances by exploiting local image region statistics, achieving simultaneous segmentation of intensity inhomogeneity images and the addition of Gaussian noise. Cai et al. \cite{CAI201879} introduced the adaptive-scale active contour model based on the maximum a posteriori framework. They constructed an adaptive scale operator and introduced a pixel membership function to achieve segmentation of severely intensity inhomogeneity images.
Li et al. \cite{LI2020443} introduced the graph space regularization of Markov Random Field to enhance the model's robustness to noise. For accurate extraction of the region of interest, Ma et al. \cite{MA2019201} proposed the Adaptive Local-Fitting-based Active Contour (ALF) method. Yang et al. \cite{SHU2023109257} presented an Adaptive Local Variances-Based Level Set Framework for segmenting intensity inhomogeneity medical images, such as cardiac MR images. Zhang et al. \cite{ZHANG2019152} proposed a multi-scale image segmentation model that simultaneously performs image denoising. Ding et al. \cite{DING2017224} utilized the Locally Based Fitting (LoG) and optimized Laplacian operator to propose the Laplacian of Gaussian model (LoGLSF). Weng et al. \cite{WENG2021115633} decomposed the observed image into three parts and proposed the Additive Bias Correction (ABC) image segmentation model. Fang et al. \cite{FANG2021397} proposed an active contour model based on a combination of mixed and locally fuzzy region energies. Wang et al. \cite{WANG2018145} defined a novel hybrid region intensity fitting energy functional. Jung et al. \cite{10.1007/s10915-016-0280-z} improved the fidelity term of traditional image segmentation models using L1 norm and proposed the Piecewise Smooth (L1PS) image segmentation model. Wu et al. \cite{WU2021126168} achieved the non-convex approximation of the Mumford-Shah (MS) model by utilizing a non-convex lp quasi-norm regularization operator and obtained segmentation results using the threshold of K-means clustering. Duan et al. \cite{9106813} established a new segmented smooth Mumford-Shah model with non-convex Lp regularization term for image labeling and segmentation. Different fitting energy functions have different effects on image segmentation models. Han et al. \cite{HAN2020107520} improved the traditional Euclidean distance and proposed a local data fitting term based on Jeffreys divergence. Liu et al. \cite{LIU201842} utilized kernel metric to enhance the model's robustness to noise. Zhi et al. \cite{ZHI2018241} proposed a hierarchical level set evolution protocol to overcome the edge leakage problem in the level set method (LSM) segmentation process.This paper proposes a segmentation model for intensity inhomogeneity and noisy images. By introducing a binary level set framework and the GL operator, this model eliminates the need for continuous reinitialization and enhances its robustness against noise. The binary level set framework replaces the traditional level set methods, thus avoiding the need for constant reinitialization. Additionally, the inclusion of the GL operator improves the model's ability to handle noisy images. By employing a three-step segmentation operator approach for solving, the effectiveness of the proposed model is demonstrated on various images.
\section{Related work}
\subsection{LBF model}
 In order to accurately segment intensity inhomogeneity images, Li et al. proposed the Locally Based Fitting (LBF) model, which utilizes the intensity information of local image regions \cite{4623242}. The LBF model introduces a variable-scale Gaussian kernel function into the data fitting term, which enables effective segmentation of intensity inhomogeneity images. Therefore, the expression of the LBF model is as follows:
\begin{equation}\label{A}
	\begin{split}
		\mathcal{M}_{\epsilon}=\mathcal{F}^{LBF}_{\epsilon}+\mu\mathcal{P}(\phi)+\nu\mathcal{L}(\phi).
	\end{split}
\end{equation}
Where $\mu$ and $\nu$ are positive parameters, $\mathcal{P}(\phi) = \frac{1}{2}\int_\Omega{(|\nabla\phi(x)-1|^2)},{\rm d}x$ is the regularization term, $\mathcal{L}(\phi) = \int_\Omega{|\nabla H(\phi(x))|},{\rm d}x$ is the length term. The data fitting function is expressed as follows:
%  其中$\mu,\nu$为正参数，$\mathcal{P}(\phi)=\frac{1}{2}\int_\Omega{(|\nabla\phi(x)-1|^2)}\,{\rm d}x$为正则项，$\mathcal{L}(\phi)=\int_\Omega{|\nabla H(\phi(x))|}\,{\rm d}x$为长度项，数据拟合函数表示如下：
\begin{equation}
	\begin{split}
		\mathcal{F}^{LBF}_{\epsilon}(\phi,f_1,f_2)=\lambda_1\int{[\int{K_{\sigma}(x-y)|I(y)-f_1(x)|^2T(\phi(y))}\,{\rm d}y]}\,{\rm d}x\\+\lambda_2\int{[\int{K_{\sigma}(x-y)|I(y)-f_2(x)|^2(1-T(\phi(y)))}\,{\rm d}y]}\,{\rm d}x
	\end{split}
\end{equation}
%其中$I:\Omega\subset R^2\to R$为图像区域，$f_1(x),f_2(x)$分别代表演化曲线$\Gamma$内外的近似灰度，$\lambda_1,\lambda_2$为正参数，$K_{\sigma}$为高斯核函数，$\mathcal{T}(x)$为Heaviside函数的平滑版本，定义为：
Where $I:\Omega\subset R^2\to R$ represents the image domain, $f_1(x)$ and $f_2(x)$ denote the approximate intensity values inside and outside the evolving curve $\Gamma$, respectively. $\lambda_1$ and $\lambda_2$ are positive parameters, $K_{\sigma}$ is a Gaussian kernel function, and $\mathcal{T}(x)$ is the smoothed version of the Heaviside function, defined as:
\begin{equation}
	\begin{split}
		\mathcal{T}_{\epsilon}(x)=\frac{1}{2}[1+\frac{2}{\pi}arctan(\frac{x}{\epsilon})]
	\end{split}
\end{equation}
%虽然LBF模型相比较于传统的CV,MS模型在分割灰度不均图像时具有优势，但是对初始轮廓敏感，且不具有分割含噪图像的能力。
Although the LBF model has advantages over traditional CV and MS models in segmenting intensity inhomogeneity images, it is sensitive to the initial contour and lacks the ability to segment noisy images.
%	\subsubsection{}

\subsection{Binary Level Set}
In order to improve the complex process of continuous reinitialization in traditional level set functions, Lie et al. proposed a binary level set model. Lie et al. suggested using a discontinuous level set function $\phi$ to represent the evolving curve $C$. The definition of $\phi$ is as follows:
%	为了改进传统水平集函数不断重新初始化的复杂过程，Lie et al 提出了二值水平集模型，Lie et al 提出使用不连续的水平集函数$\phi$来表示演化曲线$C$,$\phi$的定义如下：
\begin{equation}
	\phi(x)=\begin{cases} 
		1, &x\in inside(C)	\\
		-1, &x\in outside(C) .
	\end{cases}  
\end{equation}
%进一步的，假设分段常数$\nu$由不连续的二值水平集函数$\phi$表示，即有 $\nu=q_1$,$inside(C)$以及$\nu=q_2,outside(C)$,i.e.
Furthermore, it is assumed that the piecewise constant $\nu$ is represented by the discontinuous binary level set function $\phi$, i.e., $\nu=q_1$ when inside$(C)$ and $\nu=q_2$ when outside$(C)$.

\begin{equation}
	\begin{split}
		\nu=q_1\psi_1+q_2\psi_2,
	\end{split}
\end{equation}
\\Where $\psi_1=\frac{1}{2}(1+\phi),\psi_2=\frac{1}{2}(1-\phi)$.\\
Therefore, by combining the binary level set with the Mumford-Shah functional, the following binary level set energy functional is obtained:
%因此，将二值水平集与Mumford-Shah泛函结合，得到下述的二值水平集能量泛函：
\begin{equation}
	\begin{split}
		\mathcal{T}(q_1,q_2,\phi)=\frac{1}{2}\int_\Omega{(I-\nu)^2}\,{\rm d}x+\mu\int_\Omega{|\nabla \phi|}\,{\rm d}x
	\end{split}
\end{equation}
%其中，$\mu\ge0$,基于约束$\phi^2=1$,可以将上述的能量泛函转换为下述无约束最小化问题：
Where $\mu \geq 0$, based on the constraint $\phi^2=1$, the above energy functional can be transformed into the following unconstrained minimization problem:
\begin{equation}
	\begin{split}
		\min_{q_1,q_2,\phi}\mathcal{T}(q_1,q_2,\phi), \phi^2-1=0
	\end{split}
\end{equation}
Furthermore, by using the Lagrange projection method and augmented Lagrangian method, the aforementioned problem can be transformed into the following expression:
%进一步的，使用拉格朗日投影法以及增广拉格朗日法，将上述问题转化为以下表达式：
\begin{equation}
	\begin{split}
		\mathcal{H}_{\nu}(q_1,q_2,\phi,\lambda)=T(q_1,q_2,\phi)+\lambda\int_\Omega{(\phi^2-1)}\,{\rm d}x+\frac{1}{2}\mu\int_\Omega{(\phi^2-1)^2}\,{\rm d}x.
	\end{split}
\end{equation} 
%其中$\lambda>0$,$\mu>0$ 是拉格朗日乘子和惩罚参数。  
Where $\lambda > 0$ and $\mu > 0$ are the Lagrange multipliers and penalty parameters.

\section{The proposed model}
To correct the intensity inhomogeneity of the image and enhance the model's robustness to noise, we introduce the bias field correction $I=bc$ model into the binary level set assumption. Furthermore, the following model assumptions are proposed.
\begin{equation}
	\begin{split}
		E(c_1,c_2,\phi)=\lambda_1\int_\Omega{(I-bc_1)^2(1+\phi)^2} dx+\lambda_2\int_\Omega{(I-bc_2)^2(1-\phi)^2} dx\\+\mu\int_\Omega{|\nabla \phi|^2} dx
	\end{split}
\end{equation} 
%其中$\lambda_1,\lambda_2,\mu,\nu$是非负常数。正则项$\int_{\Omega}|\nabla \phi|^2 dx$有效抑制了水平集在演化过程中的震荡，同时增强了模型对噪声的鲁棒性。为了对上述泛函进一步求解，我们将其转换为下述的无约束最小化问题：
Where $\lambda_1$, $\lambda_2$, $\mu$, and $\nu$ are non-negative constants. The regularization term $\int_{\Omega}|\nabla \phi|^2 dx$ effectively suppresses the oscillation of the level set during the evolution process and enhances the model's robustness to noise. To further solve the above functional, we transform it into the following unconstrained minimization problem.
\begin{equation}\label{B}
	\begin{split}
		min_{c_1,c_2,\phi}E(c_1,c_2,\phi)=\lambda_1\int_\Omega{(I-bc_1)^2(1+\phi)^2} dx+\lambda_2\int_\Omega{(I-bc_2)^2(1-\phi)^2} dx\\+\mu\int_\Omega{|\nabla \phi|^2} dx+\nu\int_\Omega{(\phi^2-1)^2} dx
	\end{split}
\end{equation}
%通常会选择较大的参数$\nu$使得变量$\phi^2$与1更接近。 其中最后一项$\int_{\Omega}(\phi^2-1)^2 dx$是为了保证水平集始终维持在二值范围内。
Usually, a larger parameter $\nu$ is chosen to make the variable $\phi^2$ closer to 1. The last term, $\int_{\Omega}(\phi^2-1)^2 dx$, is included to ensure that the level set remains within the binary range.
\section{Algorithm implementation}
%为了求解问题Eq.\eqref{B},我们通过迭代最小化算法以及三步分裂算子法更新变量$c_1,c_2,b$以及变量$\phi$。因此我们将其分为以下两个部分进行求解：
To solve the problem in Eq.\eqref{B}, we update the variables $c_1$, $c_2$, $b$, and the variable $\phi$ using iterative minimization algorithms and a three-step splitting operator. Therefore, we divide it into the following two parts for solving:
(1) To solve the subproblem of $c_1$ and $c_2$, we fix the variables $\phi$ and $b$ and solve for
\begin{equation}\label{C}
	\begin{split}
		min_{c_1,c_2}E(c_1,c_2)=\lambda_1\int_\Omega{(I-bc_1)^2(1+\phi)^2} dx+\lambda_2\int_\Omega{(I-bc_2)^2(1-\phi)^2} dx
	\end{split}
\end{equation} 
where the last two terms in Eq.\eqref{B} are omitted since they are independent of the variables $c_1$ and $c_2$.
%(1)求解$c_1,c_2$子问题，我们固定变量$\phi$以及变量$b$,求解
\begin{equation}\label{C}
	\begin{split}
		min_{c_1,c_2}E(c_1,c_2)=\lambda_1\int_\Omega{(I-bc_1)^2(1+\phi)^2} dx+\lambda_2\int_\Omega{(I-bc_2)^2(1-\phi)^2} dx
	\end{split}
\end{equation} 
% 其中，\eqref{B}中的最后两项被省略，因为它们与变量$\c_1,c_2$无关。
The last two terms in \eqref{B} are omitted because they are independent of the variables $c_1$ and $c_2$.
%(2)求解$b$子问题, 我们固定变量$\phi$以及变量$c_1,c_2$,求解
(2) To solve the subproblem of $b$, we fix the variables $\phi$, $c_1$, and $c_2$, and solve for
\begin{equation}
	\begin{split}
		min_{b}E(b)=\lambda_1\int_\Omega{(I-bc_1)^2(1+\phi)^2} dx+\lambda_2\int_\Omega{(I-bc_2)^2(1-\phi)^2} dx
	\end{split}
\end{equation} 
\begin{equation}\label{D}
	\begin{split}
		min_{\phi}E(\phi)=\lambda_1\int_\Omega{(I-bc_1)^2(1+\phi)^2} dx+\lambda_2\int_\Omega{(I-bc_2)^2(1-\phi)^2} dx\\+\mu\int_{\Omega}|\nabla \phi|^2 dx+\nu\int_\Omega{(\phi^2-1)^2} dx
	\end{split}
\end{equation}
%(3)求解$\phi$子问题，固定变量$c_1,c_2,b$，求解
(3)Solve the $\phi$ sub-problem by fixing the variables $c_1, c_2, b$ and finding the solution.
\begin{equation}\label{E}
	\begin{split}
		min_{\phi}E(\phi)=\lambda_1\int_\Omega{(I-bc_1)^2(1+\phi)^2} dx+\lambda_2\int_\Omega{(I-bc_2)^2(1-\phi)^2} dx\\+\mu\int_{\Omega}|\nabla \phi|^2 dx+\nu\int_\Omega{(\phi^2-1)^2} dx
	\end{split}
\end{equation}
%通过梯度下降法，我们可以得到$c_1,c_2$的值如下所示：
By using the gradient descent method, we can obtain the values of $c_1$ and $c_2$ as follows:
\begin{equation}
	\begin{split}
		c_1=\frac{\int_\Omega{Ib(1+\phi)^2}dx}{\int_\Omega{b^2(1+\phi)^2}dx}
	\end{split}
\end{equation}
\begin{equation}
	\begin{split}
		c_2=\frac{\int_\Omega{Ib(1-\phi)^2}dx}{\int_\Omega{b^2(1-\phi)^2}dx}
	\end{split}
\end{equation}
% 固定变量$c_1,c_2,\phi$我们迭代更新变量$b$，可得下式：
By fixing the variables $c_1$, $c_2$, and $\phi$, we iteratively update the variable $b$ and obtain the following equation:
\begin{equation}
	\begin{split}
		b=\frac{\int_\Omega{\lambda_1I(1+\phi)^2c_1+\lambda_2I(1-\phi)^2c_2dx}}{\int_\Omega{\lambda_1c_1^2(1+\phi)^2+\lambda_2c^2_1(1-\phi)^2dx}}
	\end{split}
\end{equation}
%接下来，我们将通过三步分裂算子法来更新迭代变量$\phi$。因此我们先对\eqref{D}使用梯度下降法得到：
Next, we will update the iterative variable $\phi$ using the three-step splitting operator. Therefore, we first apply the gradient descent method to Eq.\eqref{E} and obtain:
\begin{equation}
	\begin{split}
		\frac{\partial\phi}{\partial t}=-\lambda_1(I-bc_1)^2(1+\phi)+\lambda_2(I-bc_2)^2(1-\phi)+\mu\nabla\phi-2\nu\phi(\phi^2-1)
	\end{split}
\end{equation}
%为了简化求解过程，我们令
To simplify the solving process, we let
\begin{equation}
	\begin{split}
		A=\lambda_1(I-bc_1)^2,B=\lambda_2(I-bc_2)^2
	\end{split}
\end{equation}
%下面，我们将使用三步分裂算子法对上述过程进行求解：
Next, we will use the three-step splitting operator method to solve the above process:
%第一步：
First step: Let $\phi^n(x)$ be the level set function computed in the nth
loop. We solve the initial value problem originating from the data
term:
\begin{equation}
	\begin{split}
		\begin{cases} 
			\frac{\partial \phi}{\partial t}=-A(1+\phi)+B(1-\phi)	\\
			\phi(x,t)|_{t=0}=\phi^n(x)
		\end{cases} 
	\end{split}
\end{equation} 
%迭代一直到获得最终的结果$\phi^{n+1,1}(x)=\phi(x,T_1),T_1>0$
The iteration continues until we obtain the final result $\phi^{n+1,1}(x) = \phi(x, T_1)$, where $T_1 > 0$.
%第二步：求解由正则项产生的非线性扩散方程：
Second step: Solving the nonlinear diffusion equation generated by the regularization term.
\begin{equation}
	\begin{split}
		\frac{\partial \phi}{\partial t}=\mu \nabla\phi
	\end{split}
\end{equation} 
%具有初值$\phi(x,t)|_t=\phi^{n+1,1}(x)$,由此我们得到第二时刻的解$\phi^{n+1,2}(x)=\phi(x,T_2),T_2>0$.
%第三步:根据初值$\phi(x,t)|_t=0=\phi^{n+1,2}(x)$,我们得到以下的非线性方程：
With the initial value $\phi(x, t)|_{t} = \phi^{n+1,1}(x)$, we obtain the solution at the second time step as $\phi^{n+1,2}(x) = \phi(x, T_2)$, where $T_2 > 0$.
The third step: Based on the initial value $\phi(x, t)|_{t} = 0 = \phi^{n+1,2}(x)$, we obtain the following nonlinear equation:
\begin{equation}
	\begin{split}
		\frac{\partial \phi}{\partial t}=-2\nu\phi(\phi^2-1)
	\end{split}
\end{equation} 
%直到某个时刻$T_3>0$,并获得第三时刻的解$\phi^{n+1}(x)=\phi(x,T_3)$.
Until a certain time $T_3 > 0$, we obtain the solution at the third time step as $\phi^{n+1}(x) = \phi(x, T_3)$.
To solve the first step, to avoid the restriction of time step, we adopt
the implicit scheme as follows:
\begin{equation}
	\begin{split}
		\frac{\phi^{n+1,1}-\phi^{n}}{\tau_1}=-A(1+\phi^{n+1,1})+B(1-\phi^{n+1,1})
	\end{split}
\end{equation} 
%$tau_1>0$是时间步长。因此我们得到第一步的结果：
$\tau_1 > 0$ is the time step. Therefore, we obtain the result of the first step as:
\begin{equation}
	\begin{split}
		\phi^{n+1,1}=\frac{\phi^n-\tau_1(A-B)}{1+\tau_1(A+B)}
	\end{split}
\end{equation} 
%在第二步中，我们对时间进一步离散得到
In the second step, we further discretize the time as follows:
\begin{equation}
	\begin{split}
		\frac{\phi^{n+1,2}-\phi^{n+1,1}}{\tau_2}=\mu\nabla\phi^{n+1,2}
	\end{split}
\end{equation} 
%其中$\tau_2>0$是时间步长。通过傅里叶快速变换，我们得到上述离散方程的解：
Where $\tau_2 > 0$ is the time step. By utilizing the Fast Fourier Transform, we obtain the solution to the discretized equation mentioned above:
\begin{equation}
	\begin{split}
		\phi^{n+1,2}=\mathscr{F}^{-1}(\frac{\mathscr{F}(\phi^{n+1,1})}{1-\mu\tau_2\mathscr{F}(\Delta)}).
	\end{split}
\end{equation} 
%其中，$\mathscr{F}$是傅里叶变换，$\mathscr{F}^{-1}$是逆傅里叶变换。
%第三步利用投影法我们得到以下解：
Where $\mathscr{F}$ denotes the Fourier transform, and $\mathscr{F}^{-1}$ denotes the inverse Fourier transform.
In the third step, utilizing the projection method, we obtain the following solution:
\begin{equation}
	\phi^{k+1}=\begin{cases}
		1&\phi^{n+1,2}\ge0\\
		-1&\phi^{n+1,2}<0.
	\end{cases}
\end{equation}
\subsection{Experiments}
This paper adopts the $JS$ and $Dice$ coefficients as evaluation metrics for image segmentation results. The parameter settings are $\lambda_1 = \lambda_2$, $\tau_1 =$. $TP$ represents true positives, which are positive samples correctly predicted by the model. $TN$ represents true negatives, which are negative samples correctly predicted by the model. $FP$ represents false positives, which are negative samples incorrectly predicted as positive by the model. $FN$ represents false negatives, which are positive samples incorrectly predicted as negative by the model. The specific definitions of $JS$ and $Dice$ coefficients are as follows:
%$TP$表示被模型预测为正的正样本，$TN$表示被模型预测为负的负样本，$FP$被模型预测为正的负样本,$FN$被模型预测为负的正样本。$JS$ and $Dice$的具体定义如下：
\begin{equation}
	\begin{split}
		Dice=\frac{2*TP}{TP+FP+TP+FN}
	\end{split}
\end{equation}
\begin{equation}
	\begin{split}
		JS=\frac{TP}{TP+FP}
	\end{split}
\end{equation} 

\begin{figure*}[!htbp]
	\centering
	\setlength{\abovecaptionskip}{-2pt}
	\subfigure{
		\begin{minipage}{0.12\linewidth}
			%	\centering
			\includegraphics[width=1.2\linewidth,height=0.1\textheight]{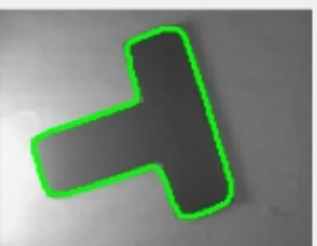}\vspace{4pt}
		\end{minipage}
	}\hspace{0.2cm}	
	\subfigure{
		\begin{minipage}{0.12\linewidth}
			%	\centering
			\includegraphics[width=1.2\linewidth,height=0.1\textheight]{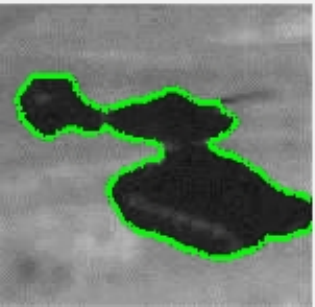}\vspace{4pt}
		\end{minipage}
	}\hspace{0.2cm}	
	\subfigure{
		\begin{minipage}{0.12\linewidth}
			%	\centering
			\includegraphics[width=1.2\linewidth,height=0.1\textheight]{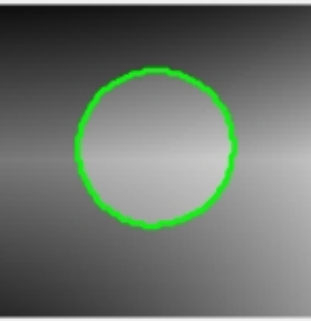}\vspace{4pt}
		\end{minipage}
	}\hspace{0.2cm}	
	\subfigure{
		\begin{minipage}{0.12\linewidth}
			%	\centering
			\includegraphics[width=1.2\linewidth,height=0.1\textheight]{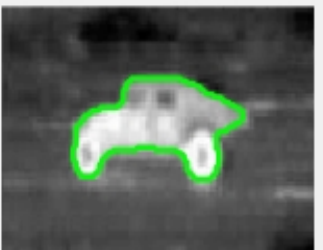}\vspace{4pt}
		\end{minipage}
	}\\
	\vspace{-10pt}	
	\setcounter{subfigure}{0}
	\subfigure{
		\begin{minipage}{0.12\linewidth}
			%	\centering
			\includegraphics[width=1.2\linewidth,height=0.1\textheight]{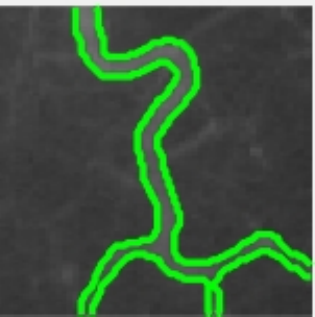}\vspace{4pt}
		\end{minipage}
	}\hspace{0.2cm}	
	\subfigure{
		\begin{minipage}{0.12\linewidth}
			%	\centering
			\includegraphics[width=1.2\linewidth,height=0.1\textheight]{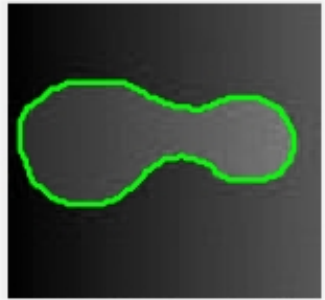}\vspace{4pt}
		\end{minipage}
	}\hspace{0.2cm}	
	\subfigure{
		\begin{minipage}{0.12\linewidth}
			%	\centering
			\includegraphics[width=1.2\linewidth,height=0.1\textheight]{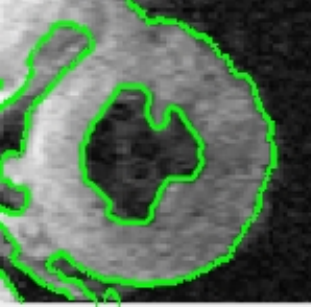}\vspace{4pt}
		\end{minipage}
	}\hspace{0.2cm}	
	\subfigure{
		\begin{minipage}{0.12\linewidth}
			%	\centering
			\includegraphics[width=1.2\linewidth,height=0.1\textheight]{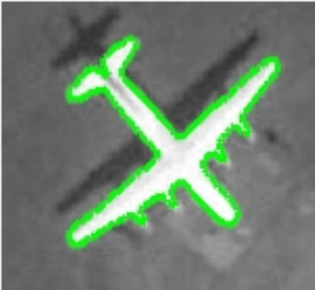}\vspace{4pt}
		\end{minipage}	
	}
	\caption{}
	\label{img9}
\end{figure*}

Regardless of whether it is low-light images, medical images, or real-world images, they often contain varying degrees of noise and intensity inhomogeneity. This requires the proposed algorithm to possess a certain robustness to images with noise and intensity inhomogeneity. To validate the segmentation performance of our model on intensity inhomogeneity, the following experiment was designed. Fig. \ref{img9} showcases the results of our proposed model in segmenting images containing bias fields. The experiments demonstrate that our model exhibits excellent accuracy in segmenting images with bias fields.

\section{Conclusion}
This paper proposes an image segmentation and bias field correction model based on binary level sets. Additionally, due to the introduction of the GL operator, the proposed model in this paper exhibits good robustness to noise and has been compared with other relevant models to demonstrate its effectiveness. Furthermore, we will further improve the proposed model's ability to handle high-noise images and optimize the modeling framework.

% Uncomment and use as the case may be
%\begin{theorem} 
%\end{theorem
	% Here goes the abstractt
	\bibliography{ref}
\end{CJK}
\end{document}